\documentclass[letterpaper, 10 pt, conference]{ieeeconf}  

\IEEEoverridecommandlockouts                              

\usepackage{times}
\usepackage{amsmath,amssymb,amsfonts}
\usepackage{graphicx}
\usepackage{wasysym}
\usepackage[ruled,vlined]{algorithm2e}
\usepackage{subfig}
\usepackage{url}

\title{\LARGE \bf
Spiral Trajectories for Building Inspection with Quadrotors
}

\author{Juan Irving Vasquez \thanks{Email address(es): contact\_author@mail.com}, David E. Troncoso Romero, Erik Zamora and  \\ Instituto Politécnico Nacional}

\author{Juan Irving Vasquez$^{1}$ and David E. Troncoso Romero$^{4}$ and Mayra Antonio-Cruz$^{2}$ and Erik Zamora $^{3}$
\thanks{$^{1}$Centro de Innovación y Desarrollo Tecnológico en Cómputo (CIDETEC), Instituto Politécnico Nacional (IPN), Ciudad de México, México.
        {\tt\small jvasquezg@ipn.mx}}%
\thanks{$^{2}$ UPIICSA, SEPI, Instituto Politécnico Nacional (IPN), Ciudad de México, México.}%
\thanks{$^{3}$ Centro de Investigación en Computación (CIC), Instituto Politécnico Nacional (IPN), Ciudad de México, México.}%
\thanks{$^{4}$ Universidad de Quintana Roo, Cancún, Quintana
Roo, México.}%
}

\begin{document}

\maketitle

\begin{abstract}
Inspection of large building is an important task since it can prevent material and human losses. A cheap and fast way to do the inspections is by sensors mounted on quadrotor vehicles. The challenge here is to compute a trajectory so that the building is completely observed while this same trajectory can be followed by the quadrotor in a smooth way. To address the problem, we propose a method that receives a 2.5D model of the target building and computes a smooth trajectory that can be followed by the quadrotor controller. The computed trajectory is a Fourier series that matches the desired behaviour. Our method has been tested in simulation and we have compared it against polynomial trajectories. Our result show that the method is efficient and can be applied to different building shapes.
\end{abstract}

\section{Introduction}

Autonomous inspection of large buildings is an important task in several areas \cite{rakha2018review}, for example, facilities examination, cultural heritage conservation, 3D modeling, etc. In general, the inspection requires to move a sensor to a set of different locations in order to observe the whole surface of a 3D structure \cite{scott2003view}. The task has been classified as a NP-problem, but by making some assumptions the problem can be solved efficiently. 


In early works, the 3D inspection of large structures have been addressed using robots attached to the surface. For instance, in \cite{siegel1998robotic}, a robotic assistant that makes contact with the surface is proposed for inspecting airplanes; or in \cite{tovakoli08} a climbing robot for inspecting pipes is proposed.  However, placing a robot over the surface is a time consuming task and could damage the surface. In consequence, the use of non invasive techniques such as vision sensors mounted on flying robots is a more suitable option. In that sense, Bircher \textit{et al.} \cite{bircher2015structural} proposed a method based on computing the views that observe all the triangles of a given 3D model. The same method refines the trajectory using an optimization criteria. Another example of inspection with flying robots, also known as micro air vehicles (MAVs), is the study of Song and Jo \cite{song2017online} where they proposed an exploration algorithm based on the motion planning technique called RRT* \cite{karaman2011sampling}. Both methods suppose that a previous complete 3D model is provided in form of a triangular mesh or point cloud. Unfortunately, such assumption is a strong limitation, since the majority of real large structures lacks of a detailed 3D model. An alternative to complete 3D models is the use of 2.5 models, where only the altitude is stored for each point in the ground. Such representations are more likely to be available for the general public because they can be obtained from blueprints or they can be approximated by carrying out photogrammetry based on fixed altitude flights \cite{remondino2011uav}. Therefore, planning inspections over 2.5D models can increase the applications of surface inspections with MAVs. 

\begin{figure}
\centering
\includegraphics[width=\linewidth]{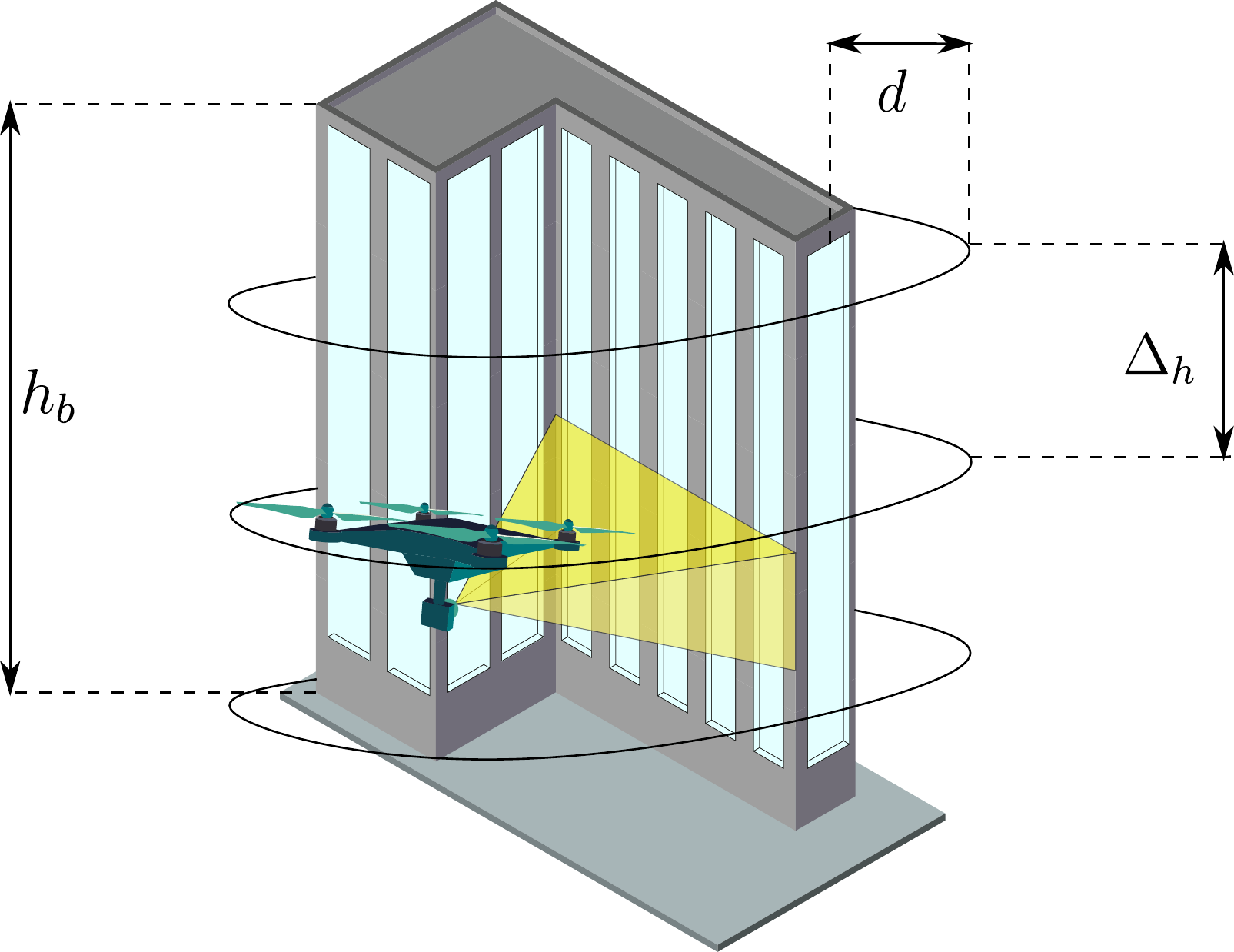}
\caption{Illustration of the inspection of a building by a quadrotor. Variables are described in the text.}
\label{fig:general}
\end{figure}

To inspect a 2.5D structure, Cheng \textit{et al.} \cite{cheng2008time} proposed to follow a sequence of horizontal circles around the target while the connection path between circles is performed by straight lines. It is worth to say that, dividing the trajectory in several paths provokes sudden changes for the controller at execution time. The problem increases when the provided trajectory does not consider the dynamics of the vehicle because the controller will be unable to follow the path. Our hypothesis is that a single smooth continuous trajectory that covers the full structure and considers the accelerations of the vehicle can improve the performance in terms of execution time and trajectory following error. 


In this paper, we propose an strategy to completely observe a large structure represented by 2.5D data with a quadrotor. The proposed strategy concatenates several motion primitives in order to completely observe the target surface. See Fig. \ref{fig:general} as an example. The motion primitives are based on Fourier series, as a result, the these trajectories are smooth and can be adjusted to satisfy the acceleration constraints of the vehicle by performing a further frequency analysis. We have also simulated the control and state estimation of a quadrotor in order to characterize the proposed method. Our results shown that the method is viable and can be applied to different building shapes.

\section{Related Work}

Building inspection has been leveraged by the use of new sensors and positioning systems like the drones which allow observations from vantage points without compromising human safety. Rakha \textit{et al.} \cite{rakha2018review} provide a general review of the parts that constitute the automated building inspection. The automation of the inspection requires to plan a path for the vehicle so that when it is followed the target surface is completely observed. Such problem is known in the literature as the coverage path planning problem \cite{de1997complete} and was initially studied in \cite{cao1988region}. In particular, when the target surface is known \textit{a priori} the problem is also known as model based planning \cite{scott2003view}.

For convenience, the related work will be classified into two kind of methods: priority on surface coverage and priority on drone agility. Surface coverage methods try to cover the surface as good as possible and optimize the path at the same time. Those methods require as input a detailed model of the target, for example, dense point clouds \cite{shi2021inspection} or triangular meshes \cite{bircher2015structural}. In general, they synthesize a set of viewpoints that senses the target and formulate a TSP problem to obtain the shortest path. For example, Bircher \textit{et al.} \cite{bircher2015structural} generate drone position by re-sampling the surfaces and optimizing the visiting order of the viewpoints. Tan \textit{et al.} \cite{tan21} uses building information modeling (BIM) for computing a set of target points at a convenient distance from the building then a genetic algorithm is applied for getting a short path. Shi \textit{et al.} \cite{shi2021inspection} use the surface normals to generate view points and then they optimize the path by modeling a sequential convex optimization problem. In some cases, both problems are optimized at the same time, as in \cite{shang2020co}.

On the other hand, methods with priority on drone agility have been less studied. They require as input a less precise model but plan robust or fast drone trajectories. For example, in \cite{sa14} a shared autonomy method is proposed, the method computes the control and stabilization with respect to a raw described target pole structure while a human user guides the inspection. In these methods, the trajectory becomes more important given that it should be easy to follow by the drone.

\section{Trajectories Generation}

Our target is to compute a trajectory so that when it is followed by the drone the 2.5D building is observed. Formally, we define a trajectory as a time function of the form:

\begin{equation}
	f(t) = (x,y,z,\dot{x},\dot{y},\dot{z}, \psi)
\label{eq:trajectory}
\end{equation} where $x$, $y$ and $z$ are the position of the MAV with respect to the inertial frame, $\dot{x}$, $\dot{y}$ and $\dot{z}$ are the translation velocities and $\psi$ is the drone yaw orientation in the same inertial frame. By defining a trajectory using equation (\ref{eq:trajectory}), inherently, we assume that the drone has a camera fixed on the front of the vehicle or that there is an additional mechanism, such as a gimbal, that can orientate the camera to the front. An example of this configuration is the Parrot Bebop Drone. We are discarding yaw and roll angles from the planning step because they can be compensated by the gimbal.

To compute the trajectory, we propose two steps. The first one generates a raw target path as a sequence of points in the workspace, then such raw path is used to generate a smooth trajectory by means of a Discrete Fourier Transform.

\begin{algorithm}[tb]
	\SetAlgoLined
	\KwData{Slices ($\mathcal{I}$), Dilation kernel ($K$)}
	\KwResult{Discrete signals, $\mathcal{X} = \{ X_l \}$ and $\mathcal{Y} = \{ Y_l \}$}
	\For{$I_l \in \mathcal{I}$}{
		$I' \leftarrow Dilate(I_l, K)$ \;
		$P \leftarrow getContour(I')$ \;
		$P \leftarrow P^T$ \;
		$X_l = \{x_i | x_i = x_{p_i}, \forall p_i \in P\}$ \;
		$Y_l = \{y_i | y_i = y_{p_i}, \forall p_i \in P\}$ \;
	}
	\caption{Generation of one dimensional discrete signals from a set of building slices.}
	\label{alg:rings}
\end{algorithm}

\subsection{Step 1: Target path}

In this first step, we construct a target path in the workspace, $W = \mathbb{R}^3$, so that it approximates the positions for the quadrotor. In general terms, this path is a set of points around the building where such points look like rings around the building, however, their shape is not necessarily a circle because it should fit the building shape. The input for this process is the building represented by a 2.5D map, namely, for each point in the ground an altitude is specified.

We start by preprocessing the input and converting it to a binary 3D grid, $M$, using ray casting \cite{amanatides1987fast}. The 3D grid is a uniform division of the 3D space like a voxelmap. In this new representation, the possible voxel labels are two: occupied and free. Then, we calculate several parameters that are needed to generate the paths accordinly to the building. First, we define an altitude increment, $\Delta_h$, as the altitude offset between two consecutive rings, so that, the target wall surface is observed and a minimum overlap is kept. This increment is computed in a similar manner than covering a 2D surface with a target given overlap \cite{vasquez2020coverage}:
\begin{equation}
	\Delta_h = \frac{d}{\mathrm{f}}h_s(1-o),
\end{equation} where $d$ is the target distance to the wall, $\mathrm{f}$ is the sensor's focal length, $h_s$ is the sensor's height and $o$ is the desired overlap. See the illustation of the variables in Fig. \ref{fig:general}. Given the altitude increment, we can calculate the number of required rings to cover whole building, $n$, as follows,
\begin{equation}
	n = \texttt{ceiling}(h_b / \Delta_h),
\end{equation} where $h_b$ is the building height. 

In the next step, we split the building into $n$ ``slices", each slice is a set of voxels at a given altitude of $M$. These slices are like cuts of the building at different altitudes. 
In the majority of cases, the number of spirals will not match the number of voxels in the axis of height because we expect a more detailed 3D grid grid with respect to the number of rings. Therefore, each building slice, $I_l$, is the set of voxels at the altitude that matches the beginning of the ring path. It is computed as follows,
\begin{equation}
	I_l = M_{i,j,k=\texttt{index}(l)},
\end{equation} where $l=1,\dots,n$, the values for the map coordinates in the $z$ axis, are computed with:
\begin{equation}
	\texttt{index}(l) = \texttt{floor}(\frac{h(l)}{h_b}r)+1,
\end{equation} where the altitude for each ring is calculated as follows,
\begin{equation}
	h(l) = \frac{h_b}{n}\left(l-1 \right). 
\end{equation}

\begin{figure}[tb]
		\subfloat[Slice from a building. Yellow pixels represent occupied space. Purple pixels represent free space. Units in pixels. \label{subfig-1:slice}]{%
			\includegraphics[trim={0 0cm 0 0cm},clip,width=0.45\linewidth]{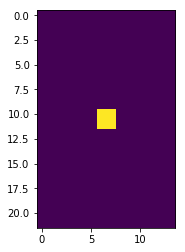}
		}
		\hfill
		\subfloat[Contour obtained after dilation and contour following. Units in pixels. \label{subfig-2:contour}]{%
			\includegraphics[trim={0 0cm 0 0cm},clip,width=0.45\linewidth]{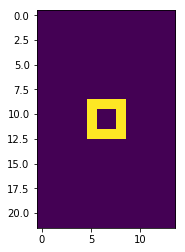}
		}
	
		\subfloat[Signal for the x component obtained from the contour. Position units in pixels. \label{fig:xpath}]{\includegraphics[width=0.45\linewidth]{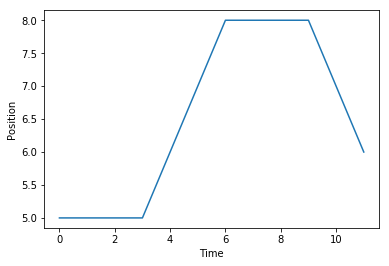}
		}
		\hfill
		\subfloat[Signal for the y component obtained from the contour. Position units in pixels. \label{fig:ypath}]{
			\includegraphics[width=0.45\linewidth]{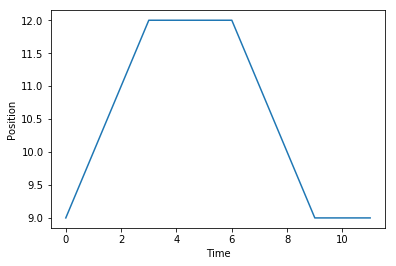}
		}
		\caption{Conversion from a building slice to a series of points that resembles a one dimensional time signal.}
		\label{fig:conversion}
\end{figure}

Once the building is divided into $n$ slices, $I_1 \dots I_n$, we process them to get a ring for each one. The process is summarized in Algorithm \ref{alg:rings}. Since the values in $M$ are only two, each slice is considered as a binary image. See Fig. \ref{subfig-1:slice} as an example. Then, for each slice, $I_l$, we dilate it using a kernel $K$, whose diameter is at least $d$. Immediately, we find the contours in the slice using Suzuki's algorithm \cite{suzuki1985topological}. Suzuki's algorithm returns all contours found, in consequence only the largest contour is kept. See Fig. \ref{subfig-2:contour}. Next, the computed contour is represented as a series of image coordinates in counter-clock-wise order, $P_l = < p_1 \dots p_m >$ where $p = (x,y)$. Then, we re-arrange the points in order to divide them into their image-horizontal and image-vertical components, $X_l$ and $Y_l$ respectively. Figures \ref{fig:xpath} and \ref{fig:ypath} show an example of the components obtained from a contour. So far, each ring represents a movement around the building at fixed height. So, to provide a movement where the drone rises at the same time, we add a set of points in the $z$ axis, $Z_l = \{z_1, \dots, z_m\}$. Since, for each ring $l$ we know the initial altitude, $h(l)$, and the final altitude $ h(l) + \Delta_h$, we use a two points line equation to interpolate all the intermediate points, so that the cardinality of all ring components will be the same, namely $|X_l| = |Y_l| = |Z_l|$. After these steps, we got three sequence of points for each ring path. For notation purposes, we concatenate all the ring components into three sets $\mathcal{X} = \{ X_l \}$, $\mathcal{Y} = \{ Y_l \}$ and $\mathcal{Z} = \{ Z_l \}$. Observing the component of each ring, we can notice that each one resembles a one dimensional discrete signal. These discrete signals describe the positions where the drone should be placed for each axis.

\subsection{Step 2: Spiral trajectory construction}

From the previous step, we have computed a discrete sequence of points for each workspace axis. These points are the desired position to be followed by the drone. However, we require a smooth and differentiable trajectory function, as presented in equation (\ref{eq:trajectory}). The reason is because its values will be used as inputs to the controller in order to navigate around the building. In addition, a differentiable function not only can provide us the velocities and accelerations required to follow the path but also they can be verified to avoid unreachable accelerations \cite{schollig2011feasiblity}. Therefore, we propose to use a Fourier series as the general form of a trajectory. One of the first works that used Fourier series for motion generation is the study of Schölig \textit{et al.} \cite{schollig2011feasiblity}, where feasible motion primitives for choreographed flights are drawn. Unlike \cite{schollig2011feasiblity}, where the trajectories are not restricted neither in space nor time, we need that the trajectories satisfy the space constraint imposed by the target ring paths ($\mathcal{X}$, $\mathcal{Y}$, $\mathcal{Z}$).

Formally, let us write the position of the drone over time as 
\begin{equation}
s(t) = (x(t),y(t),z(t)).
\end{equation}

Then, we define the general form of trajectory based on Fourier series as:
\begin{equation}
s(t) = A + \sum_{k=1}^{N} A_k \cos (\frac{2 \pi k t}{T}) + B_k \sin (\frac{2 \pi k t}{T}),
\end{equation} where $T$ is the period, $A$, $A_k$ y $B_k$ are $\mathbb{R}^3$ vectors and $N>1$. 

Based on the previous formulation, the parameters $\Omega = \{j, A, A_k, B_k, N\}$ establish the shape of the trajectory. Therefore, below we address the problem of computing a set of parameters, $\Omega$, so that a trajectory $s(t,\Omega)$ matches the target paths from the previous section. To solve the problem, we found two ways, the first one is to compute a set of Fourier components by transforming the points to the frequency domain, the second one is to optimize a set of parameters $\Omega$ so that the execution of the trajectory matches the target rings. In this paper, we follow the former approach. We apply the Discrete Fourier Transform (DFT) \cite{kumar201950} to obtain the frequencies that compose a signal that roughly matches the target points.
\begin{equation}
F_k = \sum_{t=0}^{Q-1} \mathrm{s}_t e^{-\frac{j2\pi kt}{Q} },
\end{equation} where $F_k$ is a frecuency coefficient, $s_t$ is an element of the ring path, $Q$ is the cardinality of the ring path and $j$ is the imaginary number symbol. 



\begin{figure}[tb]
    \centering
    \includegraphics[width = \linewidth]{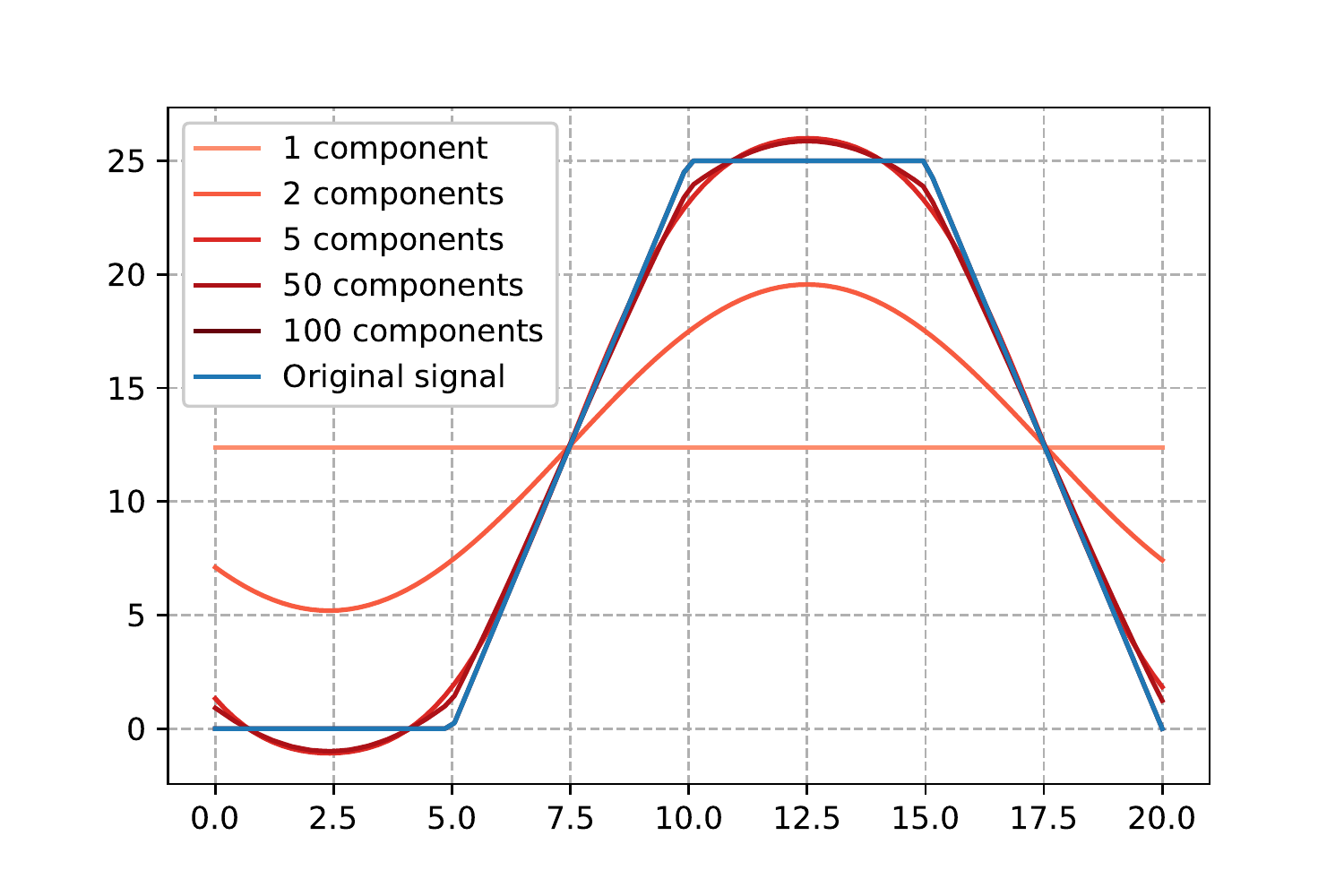}
    \caption{Approximated path using a Fourier series. The figure shows the Fourier based trajectory that approximates the target path using a different amount of Fourier components.}
    \label{fig:approximation}
\end{figure}

Then, based on the frequency representation we reconstruct the trajectory using the Inverse Fourier Transform (IFT):
\begin{equation}
	s_k = \frac{1}{Q} \sum_{t=0}^{Q-1} F_t e^{\frac{j2\pi kt}{Q} },
\end{equation}
with respect to the original path, the computed approximation has the advantage that abrupt changes are avoided. See Fig. \ref{fig:approximation} as an example of the trajectory that approximates a target path using a different amount of components.



\begin{figure}[tb]
    \centering
    \includegraphics[width = \linewidth]{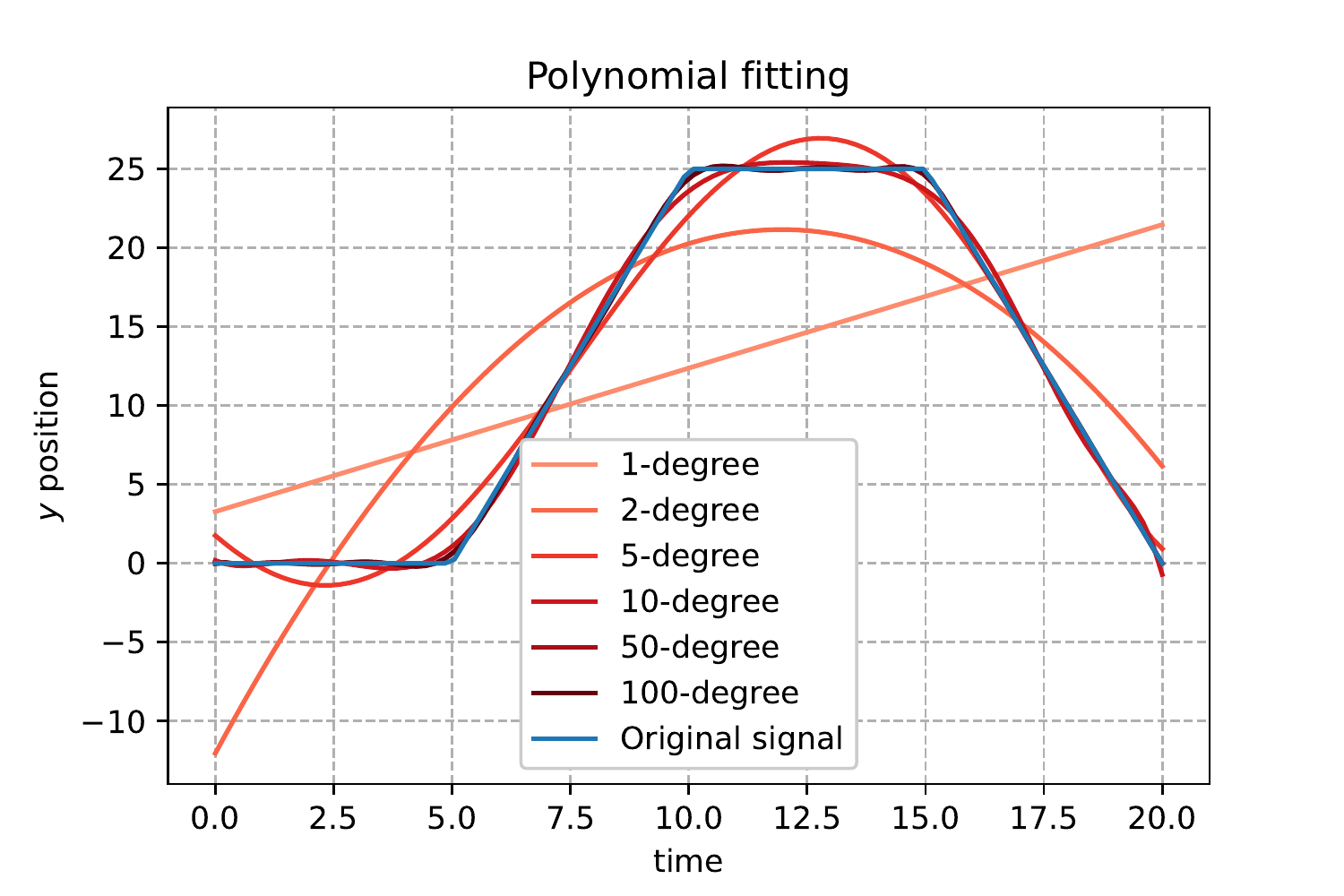}
    \caption{Approximated path polynomials.}
    \label{fig:approximation_poly}
\end{figure}

\section{Experiments}

We test and analyze the performance of the proposed method with several target ring paths. The trajectory generation system was implemented in python. For the DFT and IFT calculation their fast version was used (FFT and FIFT). The trajectory following simulation was implemented in C++ using as base the Udacity simulator \cite{dcsim}.

\subsection{Target path fitting}

\begin{figure*}[t]
		\subfloat[Square target ring path. \label{subfig:targetrp1}]{%
			\includegraphics[width=0.33\linewidth]{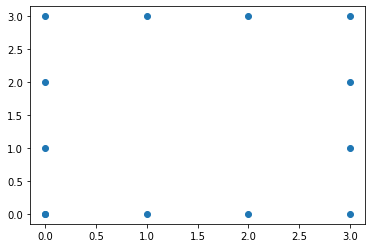}
		}
		\subfloat[Trapezoid target ring path. \label{subfig:targetrp2}]{%
			\includegraphics[width=0.33\linewidth]{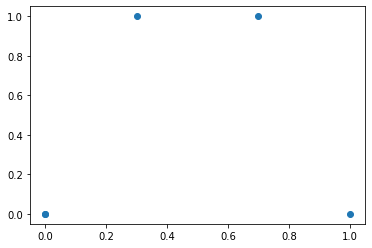}
		}
		\subfloat[Bat target ring path. \label{subfig:targetrp3}]{%
			\includegraphics[width=0.33\linewidth]{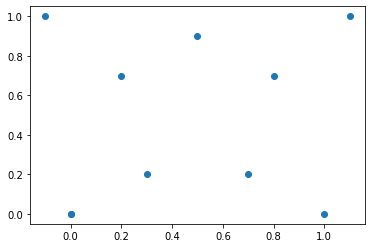}
		}
		\hfill
		\subfloat[First computed trajectory. \label{subfig:ifftp1}]{%
			\includegraphics[width=0.33\linewidth]{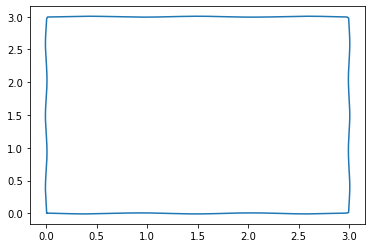}
		}
		\subfloat[Second computed trajectory \label{subfig:ifftp2}]{%
			\includegraphics[width=0.33\linewidth]{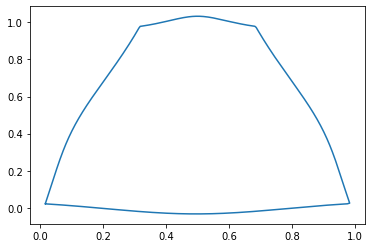}
		}
		\subfloat[Third computed trajectory \label{subfig:ifftp3}]{%
			\includegraphics[width=0.33\linewidth]{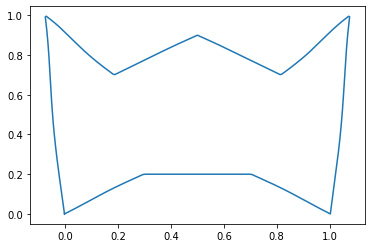}
		}
		\hfill
		\subfloat[Trajectory following. \label{fig:following1}]{
			\includegraphics[width=0.33\linewidth]{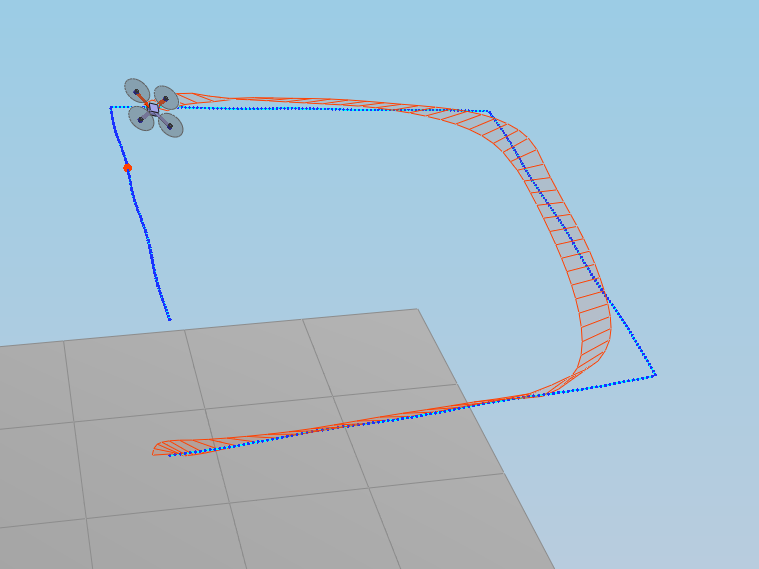}
		}
		\subfloat[Trajectory following. \label{fig:following2}]{
			\includegraphics[width=0.33\linewidth]{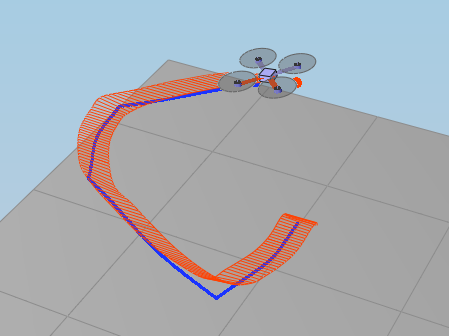}
		}
		\subfloat[Trajectory following. \label{fig:following3}]{
			\includegraphics[width=0.33\linewidth]{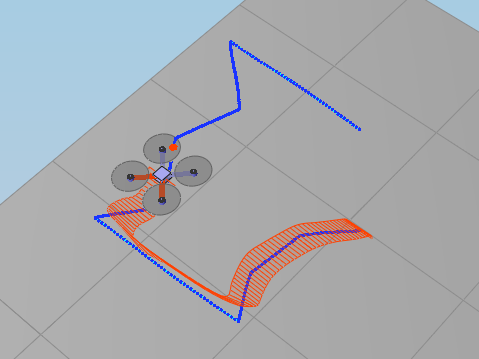}
		}
		\caption{Validation of the method for several target ring paths.}
		\label{fig:dummy}
\end{figure*}

In this experiment, we tested our method with several target ring paths and we compared its precision against a polynomial based fitting. The target paths were drawn arbitrarily and can be observed in Figs. \ref{subfig:targetrp1}, \ref{subfig:targetrp2} and \ref{subfig:targetrp3}. The figures are an square, a trapezoid and a 'bat'. Those paths were converted to one dimensional series and resampled to $N=100$ elements, in order to obtain the same number of FFT coefficients. Then the FIFT is used to get the trajectory. The rearranged and concatenated trajectories are displayed in Figs. \ref{subfig:ifftp1}, \ref{subfig:ifftp2} and \ref{subfig:ifftp3}. Those trajectories are sent to the on-board controller for being followed by the quadcopter in simulation. The trajectory followed by the quadcopter is displayed in Figs. \ref{fig:following1}, \ref{fig:following2} and \ref{fig:following3}.

In Table \ref{tab:mse} the mean squared error (MSE) between the target ring path and the generated trajectory for each approach is shown. In that table, we can observe the precision of the method by varying the number of DFT coefficients as well as the degree of the polynomial. The processing time for each experiment is in the order of milliseconds for a COLAB based computation.

Based on the experiment, we can see that both methods, Fourier based approximation and polynomial fitting, can approximate the target ring path. However, as we increment the number of components or the degree of the polynomial, the Fourier based trajectory becomes more precise. That is the case for 100 components of the FFT. 

\begin{table}[tb]
    \centering
\begin{tabular}{|l|lll|}
\hline
 & \multicolumn{3}{c|}{Figure}                                        \\ \hline
    Approach     & \multicolumn{1}{l|}{Square} & \multicolumn{1}{l|}{Trapezoid} & Bat \\ \hline
    IFT - 1 term     & \multicolumn{1}{l|}{1.05E+02}       & \multicolumn{1}{l|}{8.37E+01}          & 8.38E+01
    \\ \hline
    IFT - 2 term     & \multicolumn{1}{l|}{2.74E+01}       & \multicolumn{1}{l|}{2.31E+01}          & 2.78E+01 \\ \hline
    IFT - 5 term     & \multicolumn{1}{l|}{5.75E-01}       & \multicolumn{1}{l|}{8.81E-01}          &  5.39E+00       \\ \hline
    IFT - 50 term     & \multicolumn{1}{l|}{3.99E-01}       & \multicolumn{1}{l|}{7.25E-01}          & 2.28E+00    \\ \hline
    IFT - 100 term     & \multicolumn{1}{l|}{\textbf{8.13E-30}}       & \multicolumn{1}{l|}{\textbf{8.84E-30}}          &  \textbf{1.31E-29}    \\ \hline
    1-Deg. Poly     & \multicolumn{1}{l|}{7.65E+01}       & \multicolumn{1}{l|}{6.18E+01}          &   6.17E+01  \\ \hline
    2-Deg. Poly     & \multicolumn{1}{l|}{2.67E+01}       & \multicolumn{1}{l|}{2.01E+01}          &   2.05E+01 \\ \hline
    5-Deg. Poly     & \multicolumn{1}{l|}{1.44E+00}       & \multicolumn{1}{l|}{1.02E+00}          &  7.77E+00 \\ \hline
    50-Deg. Poly     & \multicolumn{1}{l|}{1.99E-01}       & \multicolumn{1}{l|}{1.49E-02}          & 6.71E-02 \\ \hline
    100-Deg. Poly     & \multicolumn{1}{l|}{1.36E-02}       & \multicolumn{1}{l|}{1.32E-02}          & 5.73E-02 \\ \hline
\end{tabular}
    \caption{Mean squared error for each tested approach. IFT: Inverse Fourier Transform. Deg.. Poly: Degrees of the Polynomial.}
    \label{tab:mse}
\end{table}

\section{Conclusions}

A method for computing smooth trajectories for building inspection has been presented. The method receives as input the 2.5D model of the building and computes a Fourier series trajectory. One of the advantages of the method is that sudden changes in the target path are avoided. The computation is almost in real time (in the order of milliseconds) given that the Fast Fourier Transform can be applied.

An interesting research direction is to extend the method for complete 3D models, namely, buildings with holes. In addition, an study on filtering the frequencies to smooth the trajectories at several levels could be done. Finally, in a future work, we will implement and test the proposed method in a real vehicle.

\bibliographystyle{plain}
\bibliography{inspection}

\end{document}